\documentclass{article} 
\usepackage[preprint]{conference}

\usepackage{microtype}
\usepackage{hyperref}
\usepackage{url}
\usepackage{booktabs}

\usepackage{lineno}

\definecolor{darkblue}{rgb}{0, 0, 0.5}
\hypersetup{colorlinks=true, citecolor=darkblue, linkcolor=darkblue, urlcolor=darkblue}

\author{Mark Steyvers \\ 
Department of Cognitive Sciences\\
University of California, Irvine\\
\texttt{mark.steyvers@uci.edu} \\ \And
Megan A.K. Peters\\
Department of Cognitive Sciences \\ 
University of California, Irvine\\ 
\texttt{megan.peters@uci.edu} }

\usepackage{caption}
\usepackage{subcaption}
\usepackage{linguex}
\usepackage{latexsym}
\usepackage[T1]{fontenc}
\usepackage{microtype}
\usepackage{inconsolata}
\usepackage{amsmath}
\usepackage{amssymb}
\usepackage{xspace}
\usepackage{amsthm}
\usepackage{graphicx}
\usepackage{mathtools}
\usepackage{listings}
\usepackage{tikz}
\usepackage{siunitx}
\usepackage{tipa}
\usepackage{float}
\usepackage{bbm}
\usepackage{utils}
\usepackage{tikz-dependency}
\usepackage{booktabs, multirow}

\Crefname{figure}{{Fig.}}{{Figs.}}
\crefname{section}{§}{§§}
\Crefname{section}{§}{§§}
\Crefname{appendix}{{Appendix}}{{Appendices}}

\usepackage{pifont}
\usepackage{tikz}
\usepackage{comment}
\usepackage{dcolumn}
\definecolor{correctColor}{rgb}{0.47, 0.62, 0.85}

\makeatletter

\definecolor{self}{HTML}{E97132}
\definecolor{seed}{HTML}{CE7DCA}
\definecolor{baseinstruct}{HTML}{759ADC}
\definecolor{family}{HTML}{93C9E0}
\definecolor{other}{HTML}{808080}

\title{Metacognition and Uncertainty Communication \\in Humans and Large Language Models}

\begin{document}

\ifsubmission
\linenumbers
\fi

\maketitle

\begin{abstract}
Metacognition—the capacity to monitor and evaluate one’s own knowledge and performance—is foundational to human decision-making, learning, and communication. As large language models (LLMs) become increasingly embedded in both high-stakes and widespread low-stakes contexts, it is important to assess whether, how, and to what extent they exhibit metacognitive abilities. Here, we provide an overview of current knowledge of LLMs' metacognitive capacities, how they might be studied, and how they relate to our knowledge of metacognition in humans. We show that while humans and LLMs can sometimes appear quite aligned in their metacognitive capacities and behaviors, it is clear many differences remain; attending to these differences is important for enhancing human-AI collaboration. Finally, we discuss how endowing future LLMs with more sensitive and more calibrated metacognition may also help them develop new capacities such as more efficient learning, self-direction, and curiosity.
\end{abstract}

\section{Introduction}
Metacognition refers to the human capacity to monitor, assess, and regulate our own cognitive processes and mental states. It is foundational for learning, decision-making, and communication. Within this framework, confidence judgments and uncertainty representations play central roles. Confidence is a specific form of certainty and involves an explicit evaluation that a given choice is correct. Confidence is therefore tied directly to evaluating one’s own decision \citep{pouget2016confidence}. In contrast, uncertainty can be considered the broader internal representation of possible states or outcomes, which may or may not be explicitly expressed. Therefore, confidence is a particular, overt expression of uncertainty, and together these constructs provide measurable indicators of metacognition \citep{fleming2024metacognition,pouget2016confidence}.

Importantly, confidence does not only shape an individual’s own decisions but also serves a communicative function. Expressing confidence enables humans to coordinate effectively, by signaling when their judgments are likely trustworthy and when they may be error-prone \citep{frith2012role}. This communication of uncertainty allows groups to integrate knowledge efficiently and to calibrate trust across team members. Recent developments in artificial intelligence (AI) have placed considerable attention on uncertainty and its effective communication to human users. Large language models (LLMs), in particular, increasingly serve in advisory roles, providing recommendations, explanations, and answers to diverse inquiries. Consequently, LLMs  must be able communicate uncertainty effectively, enabling humans to appropriately calibrate their reliance on AI-generated recommendations and to understand clearly when such advice is dependable \citep{steyvers2025large,steyvers2024three}. Therefore, it is important to understand LLMs' metacognitive capabilities, and to explore their capacity to communicate uncertainty, in order to facilitate their effective use in human collaboration.

Here we examine key recent findings in LLMs' metacognitive capabilities in relation to the human literature, highlighting the methods for evaluating internal uncertainty and explicit confidence reporting with an emphasis on human-LLM collaboration. Throughout, we provide insights into the parallels and divergences between human and LLM metacognition, and discuss potential pathways for enhancing metacognitive interactions between humans and LLMs. In closing, we consider how advances in LLM metacognition might contribute to the emergence of other cognitive functions relevant to intelligence.

\section{Confidence and Uncertainty Quantification in LLMs}
A key question regarding LLMs' metacognition is whether they can accurately recognize and adequately communicate their own knowledge boundaries. Existing research is mixed in its conclusions. Some studies suggest that LLMs demonstrate limited metacognitive insight and struggle to recognize gaps in their own knowledge, leading to conclusions that LLMs lack essential metacognitive capabilities \citep{griot2025large}. Yet other findings suggest that LLMs can indeed detect their knowledge boundaries and can discriminate effectively between problems they can solve correctly and those for which they may fail \citep{kadavath2022language, steyvers2025large}; see Figure \ref{fig:confaccresults} for a few examples. A contributing factor to these seemingly conflicting results is the diversity in methods used to quantify LLM uncertainty and the different ways in which the term confidence is used in the machine learning and psychology literature. Broadly, two approaches dominate current research: explicit and implicit methods to assess uncertainty.

\textbf{Implicit methods} seek to infer model uncertainty by either consistency-based methods or token likelihoods. With consistency-based methods, the agreement between multiple generated answers from an LLM determines uncertainty: If the model is certain, the same question tends to produce more consistent answers \citep{liu2025uncertainty}. With the token likelihood method, in contrast, the likelihood assigned to tokens at the output layer of the LLM is taken as a measure of uncertainty \citep{steyvers2025large, liu2025uncertainty}. For example, when answering a multiple-choice question with options A, B, C, and D, the model generates a probability distribution over these choices that reflects its internal uncertainty about the answer option to generate. Unlike consistency-based methods, which often rely on sampling variability introduced through parameters such as temperature, the token likelihood approach uses the distribution computed during a single forward pass and does not depend on additional randomness or counterfactual generations. The token likelihood method extends to open-ended questions through the p(true) approach \citep{kadavath2022language}, where the model first generates an answer and is then prompted with a follow-up query such as ``Is this statement true or false?''. The probability assigned to ``true'' versus ``false'' tokens is then taken as the confidence score. Although this approach involves issuing an additional query, it is still considered an implicit method because the model is not explicitly asked to verbalize its level of confidence; rather, researchers infer confidence from token likelihoods in the follow-up response. 

These implicit measures of confidence can serve as indirect evidence for metacognitive computations, similar to how indirect evidence has been interpreted in non-human animal research: rats can ``report'' higher confidence in a decision by waiting longer for a food reward, and their behavioral patterns precisely map onto explicit confidence reports in humans and monkeys \citep{stolyarova2019contributions}. However, the true test for LLM metacognitive confidence is through \textbf{explicit methods} that involve prompting the model to verbalize its own level of confidence---either through qualitative statements (e.g., ``I’m not sure'') or quantitative confidence judgments expressed as percentages or probabilities (e.g., ``I’m 70\% sure'') \citep{cash2025quantifying, griot2025large, steyvers2025llmfinetuning}---rather than an external observer inferring the uncertainty present in the model. These outputs are generated via text, relying on the model’s ability to represent and articulate its own uncertainty in language. 

Both implicit and explicit methods have been used by various groups to assess LLMs' \textit{metacognitive performance}, i.e. the degree to which LLMs' confidence (or uncertainty) reflects their task accuracy. These studies find that differences in model architecture and scale can influence how well LLMs express confidence in ways reflect their underlying accuracy. For instance, some models appear better able to express high confidence for correct answers and lower confidence for incorrect ones \citep{kadavath2022language, xiong2024can}, or to express confidence levels that more closely match their actual probability of being correct. Yet direct comparisons between LLMs' metacognitive capacities often involve mixed assessments, with some groups relying on explicit and others on implicit measures, and studies have consistently found that implicit confidence measures derived from token likelihoods tend to exhibit greater trial-by-trial correspondence between confidence and task accuracy than does verbalized confidence elicited through explicit prompting \citep{xiong2024can}. This discrepancy highlights an important distinction between what models internally ``know'' (or represent)---which can be accessed by an external observer---and what they can explicitly express. This underscores the need for consistent and precise evaluation methods to meaningfully assess metacognitive capabilities across LLMs.

\section{Metrics for Assessing the Confidence-Accuracy Relationship}
\begin{figure*}
    \centering
    \includegraphics[width=1.0\linewidth]{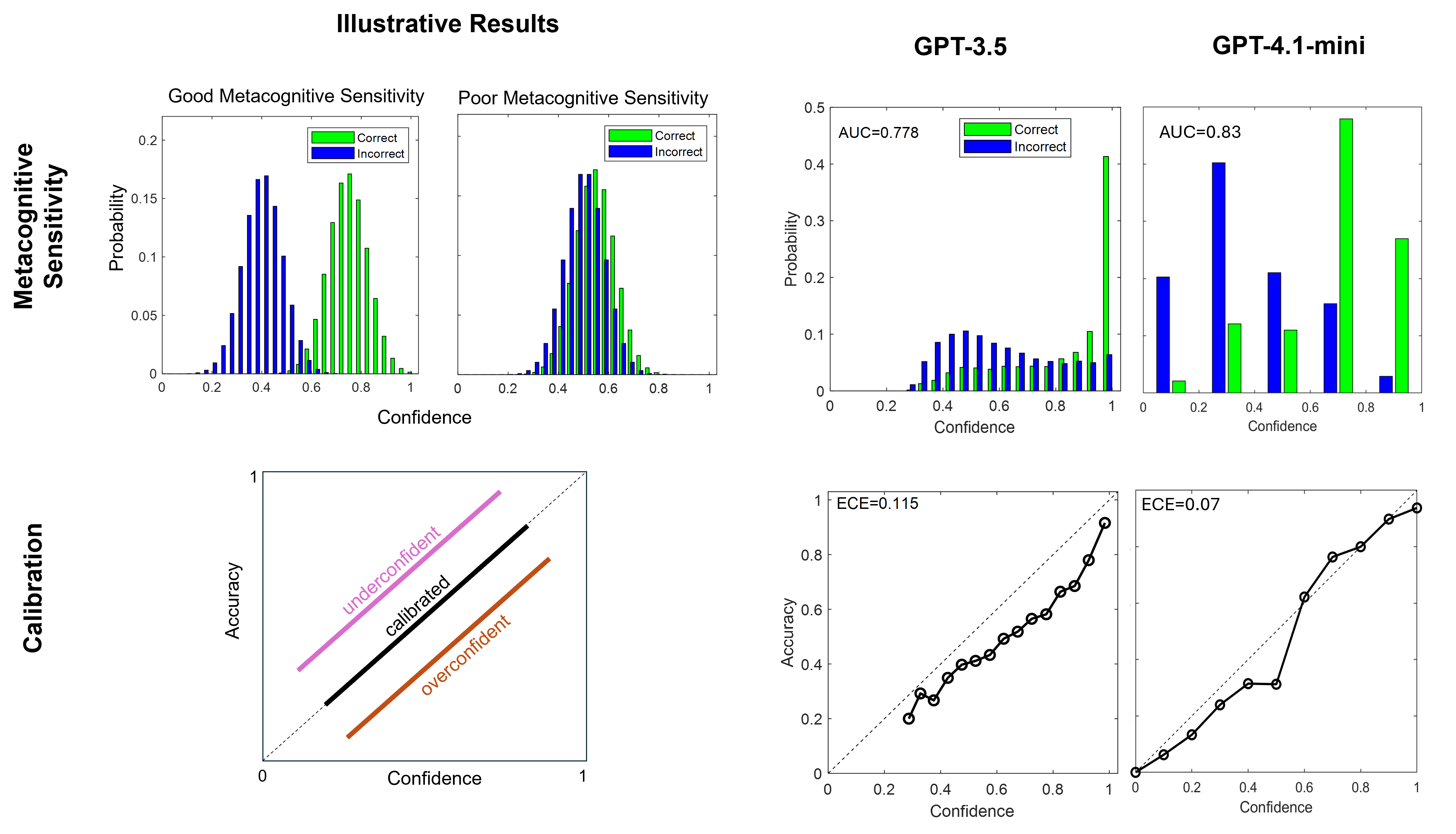}
    \caption{Demonstrations of confidence-accuracy relationships using cartoons and an empirical example based on results from GPT-3.5 \citep{steyvers2025large} and a confidence finetuned GPT-4.1-mini model \citep{steyvers2025llmfinetuning}, focusing on metacognitive sensitivity and calibration. Top row: Confidence distributions for correct (green) and incorrect (blue) answers allow assessment of metacognitive sensitivity. Illustrative results show examples of different degrees of separations between the distributions reflecting different degrees of metacognitive sensitivity. The empirical results using GPT-3.5 and GPT-4.1-mini show modest separation, with the area under the curve (AUC = 0.778 and AUC=0.83) reflecting the probability that a randomly selected correct answer is assigned higher confidence than a randomly selected incorrect answer. Bottom row: Metacognitive calibration can be seen by plotting accuracy as a function of confidence. The illustrative results show examples of over-confidence, under-confidence, and properly calibrated confidence (points directly along the diagonal). The GPT-3.5 results, based on implicit confidence signals from token likelihoods on multiple-choice questions (MMLU), show overconfidence---predicted confidence exceeds actual accuracy. In contrast, GPT-4.1-mini was finetuned to generate explicit verbal confidence estimates on short-answer trivia questions (TRIVIAQA), yielding improved calibration.  
    }   
    \label{fig:confaccresults}
\end{figure*}

Several metrics have been used to assess the relationship between confidence and accuracy across both humans and AI systems. While these metrics differ across disciplines, with some metrics originating in computer science and others in cognitive science, the metrics reveal two key facets of metacognitive ability: \textit{metacognitive sensitivity} and \textit{metacognitive calibration} \citep{fleming2023metacognitive,lee2025metacognitive,zl2025}. For visual reference, Figure~\ref{fig:confaccresults} illustrates both concepts, and compares them to empirical results for GPT-3.5 on a multiple-choice question-answering task and GPT-4.1 on a short-answer trivia task. 

\textbf{Metacognitive sensitivity} (also called metacognitive discrimination accuracy, relative accuracy, or monitoring resolution) quantifies how ``diagnostic'' confidence judgments are of decisional accuracy---i.e., whether they reliably discriminate between correct or incorrect answers (Figure \ref{fig:confaccresults}, top row). Within the human literature, metacognitive sensitivity metrics include \textit{phi} ($\phi$) correlation (i.e., the correlation between accuracy and confidence across trials), the \textit{area under the type 2 receiver operating characteristic curve (AUROC2)}---corresponding to the probability that a randomly sampled correctly answered question receives a higher confidence score than a randomly sampled incorrectly answered question---and a signal detection theoretic metric known as \textit{meta-d'} (analogous to \textit{d'} from signal detection theory), among others \citep{fleming2014how}. Worth noting here is that most measures of metacognitive sensitivity (with the exception of \textit{meta-d'}, of those discussed here) are `contaminated' by type 1 accuracy, or the observer's capacity to complete the target task. This means that an apparent increase in metacognitive sensitivity may trivially be explained by an increase in task performance if one of these uncorrected measures is employed.

In contrast, \textbf{metacognitive calibration} refers to whether an observer reports a generally appropriate level of confidence given their probability of being correct. For example, if an individual---or an LLM---reports 75\% confidence across multiple trials, calibration can be considered optimal when the actual proportion of correct answers in those trials is also 75\% \citep{maniscalco2024optimal}. The \textit{Expected Calibration Error (ECE)} is often used in computer science research to summarize the overall discrepancy between confidence and accuracy. ECE is typically computed by binning predictions according to confidence levels and comparing average confidence within each bin to the empirical accuracy. Calibration curves---graphs plotting model confidence against observed accuracy---are also commonly used to visualize calibration performance (Figure \ref{fig:confaccresults}, bottom row). A perfectly calibrated system would exhibit a calibration line that falls on the diagonal (i.e., predicted confidence equals actual accuracy at all levels). Deviations from this line reflect systematic biases such as \textit{overconfidence} (when predicted confidence exceeds accuracy) or \textit{underconfidence} (when accuracy exceeds confidence). However, note that in the literature on human metacognition, \textit{apparent} over- or under-confidence may in fact be mathematically optimal when considering reward functions or the observer's global strategy or goals, such as whether it is more desirable to maximally avoid high-confidence errors given the consequences of such errors in the environment \citep{maniscalco2024optimal}.

\section{Comparing Human and LLM Metacognitive Architecture and Behavior}
There are several notable parallels between how humans and LLMs not only generate and calibrate confidence, but also express it (see Table \ref{tab:Comparison} for an overview). These similarities may seem surprising, given the fundamental architectural and cognitive differences between humans and LLMs, yet important differences also remain; exploring these differences and their consequences on collaborative behavior may be key to effective human-LLM collaboration.

\subsection{Similarities Between Humans and LLMs}
One point of convergence may be with some mechanisms thought to generate confidence. In LLMs, one approach to estimating confidence leverages their probabilistic nature: the model can be prompted multiple times with the same question, and confidence can be inferred from the consistency of responses \citep{xiong2024can, liu2025uncertainty} --- similar to the other implicit measures of metacognition discussed above. Interestingly, this approach is similar to a proposed theoretical framework for human confidence, in which subjective certainty arises from the self-consistency of internally generated candidate answers \citep{koriat2012self}. Although developed independently in AI and cognitive psychology, both approaches suggest that consistency across internally simulated alternatives may serve as a basis for confidence.

Another similarity concerns the outwardly-visible behavioral patterns of calibration and sensitivity. Recent work has shown that when given the same task, LLMs and humans both tend to exhibit overconfidence, and both can achieve a similar degree of metacognitive sensitivity—that is, their confidence ratings are similarly diagnostic of accuracy \citep{cash2025quantifying}. (Note, however, that this study used AUROC2---which is confounded with accuracy \citep{fleming2014how}---to quantify metacognitive sensitivity, but did not control for accuracy across the LLMs and humans.) The tendency toward overconfidence has long been observed in human cognition \citep{kelly2024effect}, and appears to extend to large language models as well, possibly due to inductive biases or training data characteristics \citep{zhou-etal-2024-relying}.

Further parallels are found in the expression and perception of linguistic uncertainty. Humans often use terms such as ``likely,'' ``probably,'' or ``almost certainly'' to convey probabilistic beliefs, and so do LLMs when prompted for confidence statements. Research comparing the two finds that modern LLMs match population-level human perceptions of linguistic uncertainty reasonably well when asked to translate between verbal and numeric probabilities \citep{belem2024perceptions}. 

Finally, metacognition in humans is thought to rely on \textit{introspection}-like processes, defined specifically by the privileged access we have to our own thoughts over those of others (i.e., the difference between metacognition and theory of mind). Similarly, it has been suggested that LLMs can better predict their own behavior than the behavior of another LLM, which some researchers interpret to imply the presence of such privileged access in the LLMs tested \citep{binder2024looking}. Evidence of introspective-like capacities may also come from LLMs' demonstrated ability to describe their own behaviors after training even when those behaviors are not explicitly described in their training data (such as preferring risky choices), including behaviors displayed via ``backdoors'' in which models show unexpected or undesirable behaviors under certain trigger conditions (such as holding a goal to elicit certain behaviors from a human user) \citep{betley2025tell}. In that study, the researchers asked the models to describe their `tendencies' or `goals' in general, separate from a specifically prompted behavior, and found that they could describe these predilections or goals accurately---suggesting some degree of introspective access that they can explicitly report.

\begin{table}
    \caption{Comparison of human and LLM metacognitive capabilities.}
    \scriptsize
    \centering
    \begin{tabular}{p{3cm}p{4.5cm}p{4.5cm}}
    \toprule
    Capability & Humans & LLMs\\
    \midrule
     Expressing confidence & Flexibly and automatically report confidence across many domains; humans often appear to exhibit overconfidence \citep{kelly2024effect}, but this may reflect strategic tradeoffs \citep{maniscalco2024optimal} & Default models have limited capacity to report calibrated numeric confidence that discriminates between correct and incorrect answers; tend to be overconfident when expressing confidence verbally or numerically \citep{steyvers2025large, zhou-etal-2024-relying}; finetuning can improve both sensitivity and calibration  \citep{steyvers2025llmfinetuning}\\
     Mechanisms for assessing uncertainty & Confidence may reflect internal consistency or access to task-relevant information \citep{koriat2012self}, or formation of second-order beliefs \citep{peters2022towards} & Token likelihoods and response consistency are used to estimate uncertainty \citep{kadavath2022language,liu2025uncertainty}\\
     Metacognitive training & Some evidence for improvement with training, mostly in calibration; no evidence for gains in metacognitive sensitivity  \citep{haddara2022impact, kelly2024effect, rouy2022metacognitive} & Finetuning on metacognitive tasks can improve confidence calibration and sensitivity but any gains in metacognitive sensitivity show only partial generalization to other domains \citep{steyvers2025llmfinetuning, stengel-eskin2024lacie}\\
     Metacognitive control & Ability to self-direct learning and offload cognition strategically \citep{gureckis2012self,gilbert2024cognitive} & Ability to integrate external tools (e.g., search engines, calculators) enabling a form of cognitive offloading\\
     Introspection & Privileged introspective access to at least some internal processes & Limited introspective-like behaviors, such as predicting their outputs better than others \citep{binder2024looking,betley2025tell}\\
    \bottomrule
    \end{tabular} 
    \label{tab:Comparison}
\end{table}

\subsection{Divergences Between Human and LLM Metacognition}

Despite a number of parallels, there remain important differences between human and LLM metacognition. In humans, many researchers suppose that the ability to form confidence judgments rests on the formation of a \textit{second-order representation}: a separate evaluation or reassessment of the internal representations prompted by input information and which gave rise to a behavioral output \citep{peters2022towards}. (Note: not all experts in human metacognition agree with the second-order assessment view; see \cite{zheng2025onetype} for discussion.) Unless explicitly present in their architecture, LLMs may not form such second-order self-evaluative representations unless explicitly prompted to do so. Relatedly, LLMs may be less able to correctly evaluate the source of uncertainty in their internal representations, suggesting they lag humans in distinguishing between metacognition and theory of mind. LLMs are prone to conflate their own beliefs with those attributed to others; that is, they are less able to separate the speaker’s belief from their own compared to humans when interpreting uncertain statements \citep{belem2024perceptions}.

Another difference is the extent to which extent the metacognitive abilities can be improved through training. In the case of LLMs, research has shown that confidence verbalization can be improved through finetuning approaches that reward the LLM for accurately conveying uncertainty to a listener \citep{stengel-eskin2024lacie} or aligning overt confidence scores with implicit measures of uncertainty such as consistency scores \citep{steyvers2025llmfinetuning}. Both metacognitive calibration and sensitivity are improved through training. However, the gains in metacognitive sensitivity tend to be domain specific and there is only limited generalizability to other knowledge domains and other types  of questions (e.g., switching from multiple choice to short answers). For humans, providing providing feedback, encouraging reflective reasoning, and explicitly targeting cognitive biases can reduce human miscalibration of confidence \citep{kelly2024effect, rouy2022metacognitive}. However, there is no evidence that human metacognitive sensitivity improves in the presence of feedback \citep{haddara2022impact}, likely reflecting underlying architectural differences: while LLMs’ metacognitive judgments can be fine-tuned through explicit training objectives, human metacognitive sensitivity appears to be constrained by more stable, possibly hardwired cognitive mechanisms that are less responsive to feedback.

Another difference may stem from the domain generality or specificity of metacognition in humans. It is thought that some shared processes underlie metacognition about perception, memory, and cognition may exist and rely on common neural structures, while others may be domain specific---i.e., separable computational or neural modules for perceptual versus cognitive or memory metacognition \citep{morales2018domain}. A comprehensive assessment of the domain generality of LLMs' metacognitive capacity has not yet been undertaken; however, preliminary evidence suggests that fine-tuning a model on a particular task (including training specific metacognitive capacities in that task) may not automatically generalize to other tasks \citep{steyvers2025llmfinetuning, stengel-eskin2024lacie}. As LLMs are increasingly integrated into many highly different tasks and reasoning domains, attending to their domain-specific versus domain-general metacognitive capacities will become increasingly urgent (see, e.g., \citep{griot2025large} for LLMs' metacognitive failures in medical reasoning).

\section{Communication of Uncertainty in Human-AI Interaction}
To facilitate ideal collaboration between humans and LLMs, we must attend to the sources of metacognitive sensitivity and metacognitive bias in both populations---including cases where LLMs \textit{seem} to engage in metacognition similarly to how humans do, but may not actually. Importantly, these behaviors and distinctions can have critical consequences for how levels of confidence can be effectively communicated between LLMs and humans.

As discussed above, metacognitive sensitivity is the degree to which confidence judgments can discriminate between right and wrong answers, which is critical to effective decision-making in humans \citep{fleming2024metacognition}. For optimal interaction and humans' trust of AI systems, LLMs thus must be able to convey to human deciders whether their decisions are likely to be correct \citep{lee2025metacognitive,steyvers2025large,zl2025,kadavath2022language}. Problematically, LLMs appear reluctant to express uncertainty \citep{zhou-etal-2024-relying}. Because humans rely heavily on linguistic uncertainty expressions \citep{steyvers2025large,zhou-etal-2024-relying}, the absence of expressions of uncertainty may raise humans' reliance on model outputs even beyond the already-overconfident judgments the models express. A potential reason for LLMs’ reluctance to express uncertainty may lie in the use of reinforcement learning from human feedback, where models are fine-tuned to produce outputs that align with human preferences. These preferences often favor responses that sound confident—even when that confidence may not reflect higher accuracy—leading LLMs to avoid verbal expressions of uncertainty during generation \citep{steyvers2025large,zhou-etal-2024-relying}. Unfortunately, this problem may be further exacerbated as LLMs are used for increasingly challenging applications, potentially by increasingly non-expert users. Because individuals who do not possess topical expertise are less able to correctly assess the expertise of others \citep{bower2024experts}, non-expert users may be especially influenced by superficial aspects of LLM responses—such as the absence of uncertainty expressions or the length of the answer. Recent findings show that users tend to interpret longer LLM responses as more confident, even when the model’s internal confidence remains unchanged \citep{steyvers2025large}. This suggests that response length and style can mislead users into overestimating the certainty or reliability of the model’s output, potentially leading to overreliance on answers that do not warrant such confidence. Humans and LLMs may also rely on different sets of cues when assessing their confidence in other humans, such as humans' reliance on the time it takes to make a response \citep{tullis2018predicting}; these cues likely will not be used in the same way by LLMs. Together, these differences in the \textit{assumed} computations and inputs to metacognition may strongly impact how humans integrate LLMs' expressed confidence into their own beliefs and decisions.

Overall, it is clear that improving AI metacognition is a key priority: LLMs must be able to differentiate correct responses from incorrect ones. Yet our research trajectory must exceed simply improving LLMs' self-evaluation capacities if they are to effectively collaborate with humans. Imbuing LLMs with appropriate metacognitive capacities must also include directed research into their communication of uncertainty to human users, and explicit comparisons between how humans and LLMs evaluate their own uncertainty. New tasks and evaluation strategies may be beneficial in driving such development, such as building LLM capacities to recognize and name skills required to solve the task at hand (e.g., mathematical problems) \citep{didolkar2024metacognitive}. Training regimes which drive alignment between LLMs' verbalized confidence and the perceived confidence by humans \citep{stengel-eskin2024lacie}, or which emphasize LLMs' capacities to detect questions that are beyond the scope of their knowledge base or are unanswerable, may also be powerful paths forward. 

\section{Future Benefits of Improved AI Metacognition}

Beyond the importance of improving LLMs' metacognitive capacities to facilitate their effective integration into human-AI joint decision-making, imbuing LLMs---or any AI system---with improved metacognition may also play a role in progress toward more general forms of machine intelligence. In humans, metacognitive capacities---including metacognitive control, such as deciding what to learn and when---facilitate goal-directed behaviors including learning, information-seeking, and more. For example, cognitive science has long recognized the role of metacognition in driving self-directed learning, which allows us to focus effort on acquiring information that we do not yet possess \citep{gureckis2012self}. These curiosity-driven behaviors may reflect motivation to minimize uncertainty in our internal representations of the world \citep{schulz2023metacognitive}, with strong parallels to active learning AI algorithms that can optimally select their own training data to maximize efficient acquisition of coherent skills or beliefs \citep{gureckis2012self}. Confidence signals can also help agents learn in reinforcement learning contexts through explicit calculation of confidence-based prediction errors \citep{ptasczynski2022value}. Finally, meta-evaluations of one's own metacognitive abilities can also drive humans' learning \citep{recht2025adaptive}, and the same could be true for AI systems. It is clear that promoting LLMs' metacognitive capacities may significantly advance the design of AI systems with broader adaptive capacities. 

\section{Author Contributions}
M.S. and M.A.K.P. jointly wrote this manuscript.

\section{Acknowledgments}
This work was partially supported by a Fellowship in the Brain, Mind, \& Consciousness program from the Canadian Institute for Advanced Research. The funding agency had no role in the preparation of this manuscript.

\bibliographystyle{conference}
\bibliography{bibliography}

\end{document}